\documentclass[11pt]{article}
\usepackage{coling2020}
\usepackage{times}
\usepackage{url}
\usepackage{latexsym}

\usepackage{graphicx}
\usepackage{subfigure}

\usepackage{tabularx} 
\usepackage{placeins}
\usepackage{array}
\usepackage{multirow}

\usepackage{verbatim}
\usepackage[utf8]{inputenc}
\usepackage{makecell}

\colingfinalcopy 

\title{VICTR: Visual Information Captured Text Representation for Text-to-Image Multimodal Tasks}

\author{Soyeon Caren Han \And
 Siqu Long \And
  Siwen Luo \And
  Kunze Wang \And
  Josiah Poon 
}

\author{Soyeon Caren Han\textsuperscript{1,2}, Siqu Long\textsuperscript{1}, Siwen Luo, Kunze Wang, Josiah Poon \\
  School of Computer Science, The University of Sydney \\
  1 Cleveland Street, NSW 2006, Australia \\
  {\tt \{caren.han, slon6753, siwen.luo, kwan4418, josiah.poon\}@sydney.edu.au}\\    }

\date{}

\begin{document}

\maketitle
\begin{abstract}
Text-to-image multimodal tasks, generating/retrieving an image from a given text description, are extremely challenging tasks since raw text descriptions cover quite limited information in order to fully describe visually realistic images. We propose a new visual contextual text representation for text-to-image multimodal tasks, VICTR, which captures rich visual semantic information of objects from the text input. First, we use the text description as initial input and conduct dependency parsing to extract the syntactic structure and analyse the semantic aspect, including object quantities, to extract the scene graph. Then, we train the extracted objects, attributes, and relations in the scene graph and the corresponding geometric relation information using Graph Convolutional Networks, and it generates text representation which integrates textual and visual semantic information. The text representation is aggregated with word-level and sentence-level embedding to generate both visual contextual word and sentence representation. For the evaluation, we attached VICTR to the state-of-the-art models in text-to-image generation.VICTR is easily added to existing models and improves across both quantitative and qualitative aspects.
\end{abstract}
\blfootnote{1 Equal contribution }
\blfootnote{2 Corresponding author (Caren.Han@sydney.edu.au)}

\section{Introduction}

Over the past decade, deep learning has achieved remarkable success in multimodal research problems, such as visual question answering, visual dialog, and image captioning. However, it is quite challenging to solve the text-to-vision multimodal tasks that deal with a text input and produce a visual output, such as text to image generation, text to video generation, or text to image retrieval. Generally, natural language (incl. text) is a more natural medium for a human to describe the image that they want to generate or retrieve. However, raw text includes only limited information to fully describe and represent an image.

For example, text to image generation tasks aim to generate photo-realistic images according to the given text descriptions. The current state-of-the-art (SOTA) text-to-image generation models~\cite{xu2018attngan,zhang2017stackgan,zhu2019dm} mainly focus on generating the high resolution images by applying generative adversarial networks (GAN)~\cite{goodfellow2014generative} and rather neglect understanding input text descriptions. The most common text encoding approach in those SOTA models applies Recurrent Neural Networks (RNN) to extract global sentence-level and word-level embedding, and the last hidden state of RNN cells is used as a direct input to the GAN image generation model. However, RNN-based text encoding approaches are not capable of fully representing the rich visual semantics of input text descriptions in order to describe or generate photo-realistic visual output (e.g. image). For example, from the given sample text caption, \textit{A man on a skateboard with a brown dog outside}, the last hidden state of RNN cells holds semantics and order of words and stores words' importance in the sentence. However, those information do not include most of the required information to describe/generate the image; for example, what objects are in the image (\textit{aspect of objects})? Where are those objects (\textit{position of objects})? How to represent the relations between objects (\textit{relation between objects})? OR for example, what objects are in the image (\textit{aspect of objects}); where are those objects (\textit{position of objects}); how to represent the relations between objects (\textit{relation between objects})? 

The question is: What would be the best approach to extract the rich visual semantics of input text descriptions in order to describe/generate an image? In this paper, we introduce a new Visual Contextual Text Representation (\textit{VICTR: Visual Information Captured Text Representation}), which represents the input text description with its visual semantics, as shown in Figure \ref{fig:architecture}. The proposed model, VICTR, can be applied to diverse text-to-image multimodal tasks.

VICTR has five different modules: 1) Text to Scene Graph Parsing, 2) Scene Graph Embedding, 3) Positional Graph Embedding, 4) Visual Semantic embedding, 5) Visual Contextual Text Representation. First, we extract scene graphs from input text descriptions in order to define what objects, attributes and relations should be in the image. The scene graph is initially proposed by \newcite{johnson2015image} in order to represent the objects, its attributes and their relations in the image. Inspired by this, we generate scene graphs from the raw text description by using dependency parsing and transformer-based object-attribute-relation classification. Then, we train the extracted object, attribute and relation nodes via Graph Convolutional Networks (GCN) in order to generate the \textit{visual contextual text representation}, a word-level representation which incorporates textual syntactic and visual semantic information. Finally, it aggregates with word-level and sentence-level embedding respectively in order to generate a visual contextual word representation and visual contextual sentence representation.

For evaluating quantitative and qualitative aspects of our proposed model, we attach VICTR to the SOTA models in text-to-image generation. Thorough experiments on the COCO benchmark dataset demonstrate the superiority of VICTR with respect to both semantic consistency and visual reality. The main contribution is summarised in the end of the Sec. \ref{section:maincon}

\section{Related Works and Contributions} 
We explore research trends in diverse text-to-vision multimodal tasks, which use text information as an only input to produce a visual output, including text to image generation and text to video generation.

\subsection{Text to Image Generation}
From 2016, text to image generation tasks have been explored by applying conditional GANs 
\cite{reed2016generative,zhang2017stackgan,zhang2018photographic} with the text caption as input. AttnGAN \cite{xu2018attngan} is the first method that utilized a word-level embedding and fused it with image vectors in an attention mechanism to identify the contributing words of sub-regions in the generated images. Extended from AttnGAN, MirrorGAN \cite{qiao2019mirrorgan} applied the same attention mechanism on both sentence and word embedding to capture the global semantic consistency between generated images and input texts. SEGAN \cite{tan2019semantics} proposed an adaptive attention mechanism on word-level embeddings to ensure only relevant words on the generated images would obtain attention-weight. SD-GAN \cite{yin2019semantics} with a Siamese structure is utilized to guarantee the semantic alignment between generated images and captions. DM-GAN \cite{zhu2019dm} proposed a dynamic memory network to fuse word embedding and image representations for image generation. Most SOTA models applied bidirectional LSTM (RNN)-based text encoding, which contains only information that represents the order of the words and the words' importance in the given text caption. The output of the RNN-based text encoding is not enough to represent rich visual semantics in order to directly generate the image, hence, the images from current SOTA models are not successful in aligning with the given text description.

\subsection{Text to Video Generation}
Similar to text to image tasks, most text to video generation models learn the caption via RNN cells as conditional input representation to be used with Variational Autoencoders (VAE) or GANs in order to generate image frames. The main difference is that video generation considers how to model the temporal dependency by conditioning on the corresponding text captions. Sync-DRAW \cite{mittal2017sync} applies the recurrent attention-based VAE to create a temporally dependent sequence of frames but it still applies LSTM for input text encoding. GAN-based text to video generation approaches also apply RNN-based encoding to handle the input text captions. TGANs-C \cite{pan2017create} utilises GANs with three discriminators to generate video based on input text captions encoded by Bi-LSTM. IRC-GAN \cite{deng2019irc} proposed a Mutual-information Introspection (MI) that measures the semantic similarity between text and generated video through a two-stage process. The conditional text input is represented through Bi-LSTM network. TFGAN \cite{balaji2019conditional} applies a scheme via generating discriminative convolutional filters from text features and then convolves them with image features in the discriminator. It applies a CNN (Convolutional neural network)-based text encoder but it still does not represent the sufficient visual semantics from text captions in order to generate the video.

\subsection{Main contribution} \label{section:maincon}
Most existing models for text-to-image and text-to-video generation tend to have a RNN or CNN-based sentence feature from the raw text for modeling the cross-model relation with the generated visual output. Hence, we now present our model, VICTR, the successful approach to extract the rich visual semantics of input text descriptions in order to describe/generate an image.
The proposed VICTR is evaluated with text-to-image generation tasks. The model with VICTR outperforms the performance of original SOTA models in photo-realistic image generation based on text input. The major contributions of this work are summarised as follows:
1) The paper provides an example of capturing rich visual semantic and geometric relation information from raw text input. 
2) The paper proposes a new visual information captured text representation for text-to-image generation tasks, which has not been reported before. The proposed text representation model can be usable with any text-to-vision multimodal tasks.

\begin{figure*}[t]
\centering
  \includegraphics[width=1\textwidth]{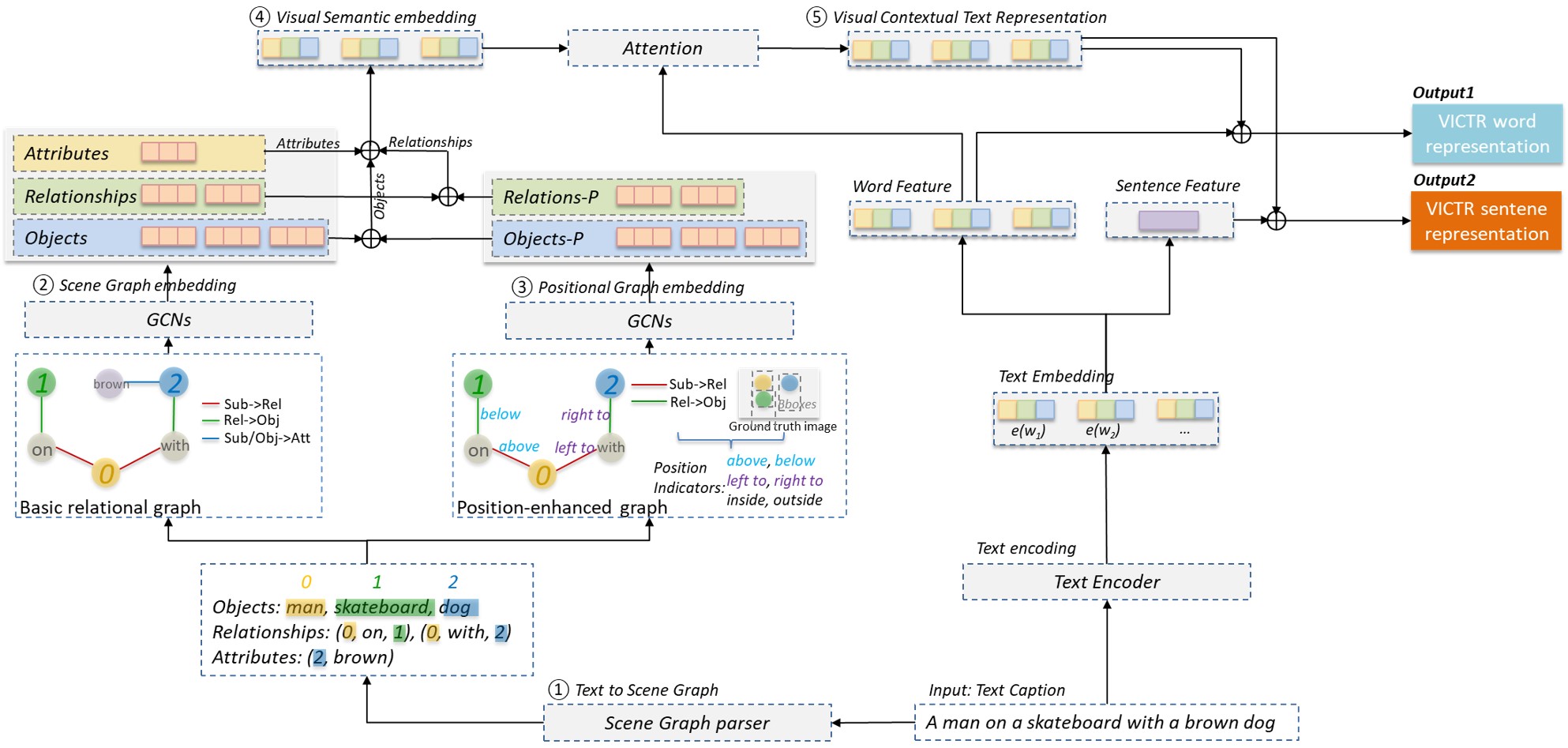}
  \caption{Schematic of the proposed visual contextual text representation (VICTR)}
  \label{fig:architecture}
\end{figure*}

\section{Methodology}
As shown in Figure \ref{fig:architecture}, the proposed visual contextual text representation, VICTR (Visual Information Captured Text Representation), mainly focuses on capturing and representing the visual semantic information (i.e. \textit{location or aspect of the object in the image, positional relation between objects}) from raw text descriptions. This is crucial for text-to-image generation tasks. In summary, the architecture of the proposed VICTR is composed into five modules: 1) Text to Scene Graph Parsing, 2) Scene Graph Embedding, 3) Positional Graph Embedding, 4) Visual Semantic Embedding, 5) Visual Contextual Text Representation. Note that the first module \textit{text to scene graph} can be considered as a pre-processing step to generate visual information embedded text representation for the rest of the architecture.

\subsection{Text to Scene Graph Parsing}
Based on the given raw text description, including image caption or scene description, we firstly extract a graph-based semantic representation, called scene-graph \cite{johnson2015image}, which explicitly represents object instances, their attributes, and the relation between objects. This simple graph representation describes visual scenes/images in great detail. Inspired by this idea, we generate scene graphs based on input text descriptions. Like the nature of text-to-image generation tasks, we use only text descriptions in order to extract the scene graph with rich visual semantics (\textit{objects, attributes, relations of the image}). In order to parse scene graphs from the given text caption, we firstly recognise the syntactic structure of the text descriptions by applying a universal dependency parser; in this research, we applied the Stanford enhanced dependency Parser \cite{chen-manning-2014-fast}. However, the output of a dependency parser would not be enough to directly represent the number of objects (as well as its attributes and relations between objects) that should be drawn in the scene graph. Hence, we have a semantic enhancement processing component, \textit{quantity checker}. The quantity checker aims to detect the number of objects that the scene graphs need to include. For example, the following two text captions \textit{two men are riding brown horses} and \textit{two men are riding a brown horse} include different semantic information: the former would have two \textit{man} objects and two brown \textit{horse} objects but the latter could contain two \textit{man} objects and one brown \textit{horse} object. We duplicate the individual nodes in the dependency graph according to the value of their quantificational modifiers. In addition to this, we also cover some quantificational determiners by using the quantifier expression rule list, such as \textit{both of},  \textit{a dozen of}, or \textit{a lot of}. From the  syntactic and semantic integrated graph, we extract all nouns to classify into object classes, and retrieve all adjectives as attribute types of the specific object (pairwise classification). The relation between objects is detected if the word is the predicate or preposition of two different objects.

As a result, each text caption of an image can derive one scene graph $G\left(O, R, A\right)$, where $O=\left\{o_{1}, o_{2}, ..., o_{n}\right\}$ is a set of objects, $R=\left\{r_{1}, r_{2}, ..., r_{m}\right\}$ is a set of relations, and $A=\left\{a_{1}, a_{2}, ..., a_{k}\right\}$ is a set of attributes.  Especially, each object ${o_{i} \in O}$ is assigned with a super-class\footnote{Super-classes are from the COCO or Mediawiki(https://wiki.dbpedia.org/) can be extracted the parent class of each object} $t\in T=\left \{ t_{1},t_{2},...,t_{\mu } \right \} $. 

\subsection{Scene Graph Embedding}
We convert the extracted scene graph to a vectorised graph representation to produce useful feature representations of nodes and edges in the object-attributes-relation networks. We apply GCNs to model the relative nearness of nodes and edges in the scene graph and preserve the visual semantics.

The basic relational graph $G_{b}$ represents visual semantic alignment between object and relation as well as object and attribute in scene graphs. We train the graph using GCNs to produce scene graph embeddings. As shown in Figure \ref{fig:architecture}, each object $o_{i}\in O$, relation $r_{i}\in R$, and attribute $a_{i}\in A$ is made as a node of the graph $G_{b}$. Then the object-to-relation connection and relation-to-object connection are represented as edges $e_{o_{i}\to r_{j}}$ and edges $e_{r_{j}\to o_{i}}$ respectively. Similarly, edge $e_{o_{i}\to a_{t}}$ indicates the object-attribute alignment. For edge $e_{o_{i}\to r_{j}}$, $e_{r_{j}\to o_{i}}$, and $e_{o_{i}\to a_{t}}$, the weight is calculated based on the equations:
\begin{equation}
    W_{e_{o_{i}\to r_{j}}} = \frac{\textit{number of}\ e_{o_{i}\to r_{j}}}{\textit{number of}\ e_{o_{i}\to R}}
\end{equation}
\begin{equation}
    W_{e_{r_{j}\to o_{i}}} = \frac{\textit{number of}\ e_{r_{j}\to o_{i}}}{\textit{number of}\ e_{r_{j}\to O}}
\end{equation}
\begin{equation}
    W_{e_{o_{i}\to a_{t}}} = \frac{\textit{number of}\ e_{a_{t}\to o_{i}}}{\textit{number of}\ e_{a_{t}\to O}}
\end{equation}
The edge weight to the node itself would be 1. The edge weights are compiled into an adjacency matrix combined with the graph degree matrix and are passed into a 2-layer GCN to be trained through mapping each object to its corresponding super-class. We denote node embeddings for an object, an attribute and a relation as $EB_{o},EB_{a},EB_{r}\in R^{B}$.

\subsection{Positional Graph Embedding}
In section 3.2, the scene graph-based basic relational graphs mainly focus on the semantic relations (predicates or preposition) between objects, e.g. \textit{ride} from the text description \textit{man rides a horse}. It provides the lingual semantics of objects and relations but it is not still enough to fully describe the image with geographical information, such as location of objects or the relative position (e.g. \textit{left to}) between objects, which includes an indicative and explicit location of one object in relation to another. Hence, we propose a position-enhanced relational graph $G_{p}$ that focuses on visual semantics of relations between objects. Six relative geometric relations are chosen and denoted as $p \in$  \{\textit{left of}, \textit{right of}, \textit{above}, \textit{below}, \textit{inside}, \textit{surrounding}\} to represent edges $e_{o_{i}\to r_{j}}$ and edges $e_{r_{j}\to o_{i}}$. To train these edges, the geometric relation is detected based on the gap between bounding boxes of one to another object. Considering that one object may correspond to multiple geometric relations, we generate individual graphs for each type of geometric relations. The weights of edges $e_{o_{i}\to r_{j}}$ and edges $e_{r_{j}\to o_{i}}$ in a graph $G_{p}$ of six geometric indicators are calculated as those in the basic relational graph. For each graph, the edge weights are compiled into an adjacency matrix combined with the graph degree matrix, and passed into a 2-layer GCN to train each object with its corresponding super-class. The object-level and relation-level node embedding in each of the six position-enhanced relational graphs are concatenated and produces the positional object-level node embedding $EP_{o}\in R^{P}$ and relation-level node embedding $EP_{r}\in R^{P}$.

\subsection{Visual Semantic Embedding} 
We now integrate object, relation, attribute-level embeddings from basic relational graphs and position enhanced graphs in order to produce the comprehensive visual semantic embedding for scene graphs. The visual semantic embedding is composed to three aspects: 1) Object-level embedding $E_{o}\in R^{B+P}$: concatenate $EB_{o}$ and $EP_{o}$ of object $o_{i}\in O$, 2) Relation-level embedding $E_{r}\in R^{B+P}$: combine $EB_{r}$ and $EP_{r}$ for each relation $r_{j}\in R$, 3) Attribute-level embedding $E_{a}=EB_{a}\in R^{B}$. For each object in one scene graph, the object embedding is concatenated with its attribute embedding as well as the corresponding relation embedding to produce the final visual semantic embedding $E_{vs}\in R^{2\ast \left ( B+P \right )+B}$.

\begin{figure}[t]
\centering
\subfigure[animals and electronics]{
\includegraphics[width=0.25\textwidth]{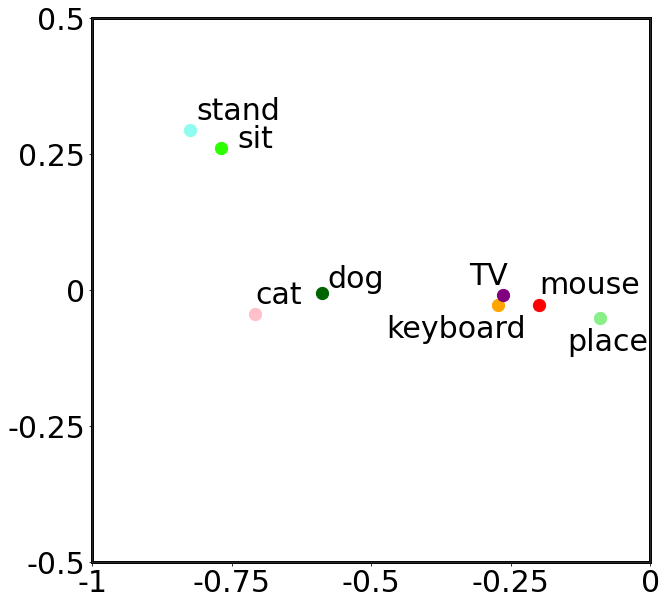}
}
\label{fig:subfig2}
\subfigure[kitchen-wares and vehicles]{
\includegraphics[width=0.25\textwidth]{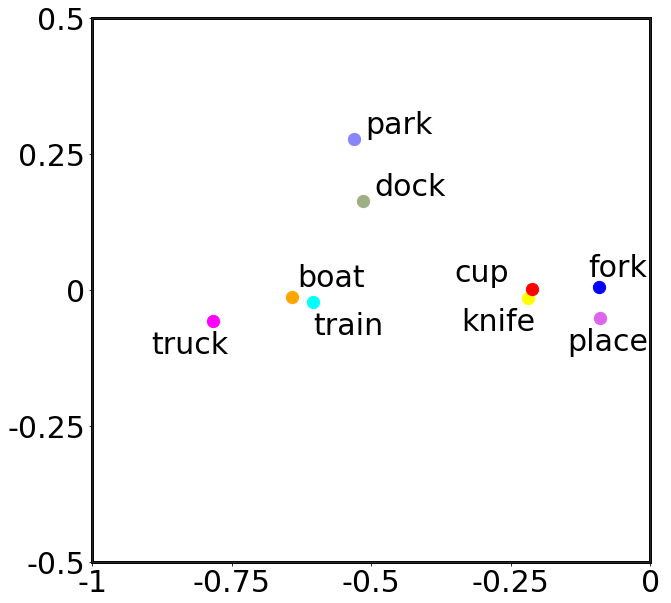}
}
\subfigure[foods and sports]{
\includegraphics[width=0.25\textwidth]{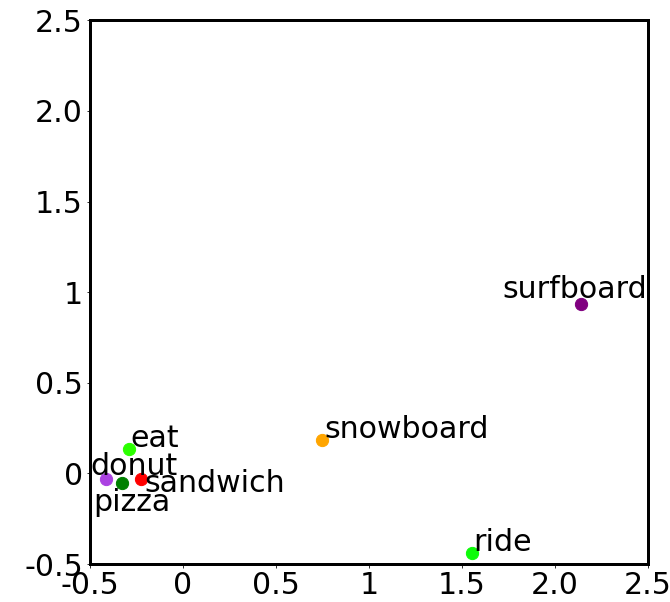}
}

\label{fig:subfig3}

\caption[]{Two-dimensional PCA projection of 1200-dimensional visual semantic vectors of objects and 500-dimensional position-enhanced vectors of relations. The figures illustrate ability of the model to automatically organise concepts and learn implicitly the similarity between them, as during the training we did not provide any supervised information but only the text description of images in COCO2014.}
\label{fig:embed_plot}
\end{figure}

Based on the produced final visual semantic embedding, we now visualise the ability of the proposed embedding model to automatically organise different aspects of objects and learn implicitly the relations between them. Figure \ref{fig:embed_plot} illustrates the visual semantic vectors of diverse objects and the position-enhanced relation vectors that appear frequently with those objects from the COCO2014. In Figure \ref{fig:embed_plot}(a), animal objects \textit{cat}, \textit{dog} are close to each other while being far away from the electronics objects, such as \textit{mouse} and \textit{TV}. These electronics objects are close to the relation \textit{place} because it is commonly used with them instead of the relation \textit{sit} or \textit{stand}. Similarly, the relations \textit{park}, \textit{dock} are close to the group of vehicle objects \textit{truck}, \textit{boat}, \textit{train} but far from the kitchenware object cluster in Figure\ref{fig:embed_plot}(b). This pattern can be also found in Figure\ref{fig:embed_plot}(c) as it is shown that the food objects are gathered together.

\subsection{Visual Contextual Text Representation}
The proposed visual semantic embedding strongly integrates the semantic information of an object with its attributes and the relations attached to it as well as the positional (geometric) relations between the object and others. In order to seamlessly grain this visual semantic information into the text representation, we integrate it into the word and sentence representation from raw text using attention mechanism. The attention mechanism is applied between the $E_{vs}$ and the corresponding text word embedding $E_{word}\in R^{L\times D}$, which is derived from the text encoder with the sequence length $L$ and the dimension of word embedding $D$. The attention is inspired by \cite{vaswani2017attention}, and we made $E_{vs}$ as both $K$ and $V$ while taking $E_{word}$ as $Q$:
\begin{equation}
\textup{Attention}\left ( Q(E_{word}), K(E_{vs}), V(E_{vs}) \right ) \\= 
    \textup{softmax}\left (  (W^{T}E_{word})E_{vs}^{T} \right )E_{vs}
\end{equation}
Here $W$ is a learnable weight to map the word representation to the visual semantics space. The attended visual semantic embedding $E_{vs'}\in R^{L\times 2\ast \left ( B+P \right )+B}$ represents the importance of each object-based visual semantic information to each word in the sequence of a text caption. The $E_{vs'}$ is concatenated with the word embedding $E_{word}$ to get the visual contextual word representation $E_{VICTR-W}$. Similarly, the object-information over all the words are summed up via $E_{vs'}$, and then concatenated with the sentence embedding $E_{sent} \in R^{D}$ derived from the text encoder to get the visual contextual sentence representation $E_{VICTR-S}$.

\section{Evaluation Setup}
\textbf{Baselines}
Three text-to-image generation SOTA models, StackGAN~\cite{zhang2017stackgan}, AttnGAN~\cite{xu2018attngan}, and DM-GAN~\cite{zhu2019dm} were selected as baselines, which all used text representation as the only source for image generation. We replaced the original text representation with our proposed VICTR and compared the generated images.

\noindent\textbf{Dataset}
We evaluated the model performance on COCO2014~\cite{lin2014microsoft}\footnote{COCO2014 \url{http://cocodataset.org/}}, which is the most common benchmark and contains photo-realistic images with diverse objects and relations. Detailed dataset statistics are shown in Table~\ref{tab:dataset_table}. Each image has 5 corresponding image descriptions and we selected the caption which generates the richest scene graph. We used bounding box features to train the geometric relations of multiple objects.

\begin{table}[h]
    \centering
    \small
    \begin{tabular}{|c|c|c|c|}
    \hline \bf Datasets & \bf Images & \bf Captions & \bf Objects \\ \hline
    Train & 82,783 & 413,915 & 604,907  \\\hline
    Test &40,504 & 202,520 & 291,875 \\
    \hline
    \end{tabular}
    \caption{\label{tab:dataset_table} COCO2014 Statistics ~\cite{lin2014microsoft}}
\end{table}

\noindent \textbf{Implementation Details} We used 2-layer GCN with the hidden layer of dimension 200 and 50 for each basic relational graph and position-enhanced graph, and trained the GCNs with learning rate 0.02 for 200 epoch. The number of trainable parameters for the basic relational graph GCN is 17,258,800 while for the six position-enhanced GCN are 492,200 (left of), 488,400 (right of), 632,000 (above), 657,200 (below), 63,800 (inside) and 54,600 (surrounding) respectively. The final hidden state is used as embedding and the concatenated $E_{vs}$ is 1200d. For AttnGAN and DM-GAN, we used their word feature $E_{word}$ (256d) and the sentence feature $E_{sent}$ (256d) extracted by the pre-trained DAMSM text encoder and applied the attention with $E_{vs}$. We replaced the derived $E_{VICTR-W}$ and $E_{VICTR-S}$ with the initial word feature $E_{word}$ and sentence feature $E_{sent}$ respectively in the original models. We kept all the best-performance configuration as it is from the three original models for comparison. We trained the StackGAN with VICTR on stage-I for 600 epoch as the same in the stackGAN paper and trained the StackGAN with VICTR on stage-II and the other two models (AttnGAN VICTR and DM-GAN VICTR) for 150 epoch. StackGAN-VICTR has the same number of trainable parameters as StackGAN, which are 32,735,457 for stage-I and 197,327,218 for stage-II. For AttnGAN-VICTR and DM-GAN-VICTR, there are 256,168,838 and 44,154,260 respectively for the GAN part and 614,400 for the weight $W$. All experiments are conducted on 24 GB NVIDIA TITAN RTX GPU with 10.2 CUDA. With our environment, it took around 24(21)min/epoch on stage-I and 45(40)min/epoch on stage-II training for StackGAN with VICTR (StackGAN), 57(51)min/epoch for AttnGAN with VICTR (AttnGAN) and 60(56)min/epoch for DM-GAN with VICTR (DM-GAN).

\noindent \textbf{Evaluation Metrics} We use Inception Score (IS), Frech\'et Inception Distance (FID) and R-precision to quantitatively evaluate the model performance on 30,000 generated images\footnote{The code of three quantitative evaluation metrics are from: \url{https://github.com/MinfengZhu/DM-GAN}}. IS \cite{salimans2016improved} uses Kullback-Leiber (KL) divergence to compare the similarity between each generated image label probability distribution and the marginal probability distribution of all generated images, the higher the IS, the better the model is to generate diverse and distinct images. FID \cite{heusel2017gans} is an improved version of IS, comparing the Frech\'et distance between the maximum entropy distribution of the generated images and the real images. The lower the FID, the more similar the generated images to the real images. R-precision measures the consistency between the generated image and the input text. We followed \newcite{xu2018attngan} and set $R=1$, comparing the cosine similarity between generated image vector and input text embedding to find the top $r$ captions that are relevant to images and calculate R-precision as $r/R$. The final score is taken as the average of R-precision of all images, the higher the score, the better consistency between generated images and captions.

\begin{table}[t]
    \centering
    \small
    \begin{tabular}{|p{3.5cm}|p{1.8cm}|p{1.8cm}|p{1.8cm}|}
    \hline 
    \bf{Methods} & \bf{IS}$\uparrow$ & \bf{FID}$\downarrow$ & \bf{R-precision}$\uparrow$ \\
    \hline
    StackGAN & 8.45~$\pm$~.03& - & - \\
        \bf{StackGAN with VICTR} & \bf{10.38~$\pm$~.20} & - & - \\
    \Xhline{1pt}
    AttnGAN & 25.89~$\pm$~.47& 35.49& 85.47~$\pm$~3.69 \\
        \bf{AttnGAN with VICTR}  & \bf{28.18~$\pm$~.51} & \bf{29.26}& \bf{86.39~$\pm$~.0039} \\
    \Xhline{1pt}
    DM-GAN & 30.49~$\pm$~.57& 32.64& 88.56~$\pm$~0.28\\
        \bf{DM-GAN with VICTR} & \bf{32.37~$\pm$~.31} & \bf{32.37}& \bf{90.37~$\pm$~.0063} \\
    \hline
    \end{tabular}
    \caption{\label{tab:result_table} Inception score (IS), Fréchet Inception Distance (FID), R-precision results of VICTR-based models comparing to all baseline models.}
    \vspace{-4mm}
\end{table}

\begin{figure*}[t]
\centering
  \subfigure[StackGAN VS StackGAN-VICTOR]{
\includegraphics[width=0.31\textwidth]{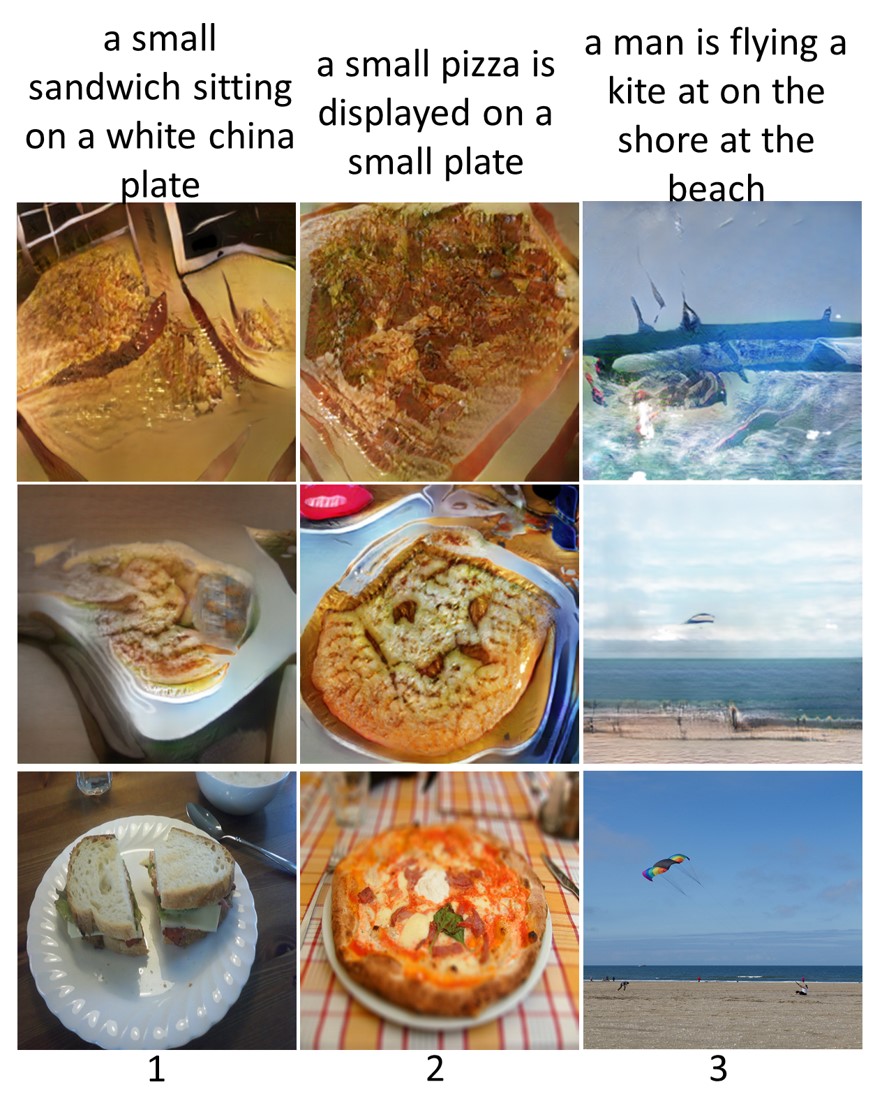}
}
\label{fig:comparision_a}
\subfigure[AttnGAN VS AttnGAN-VICTOR]{
\includegraphics[width=0.31\textwidth]{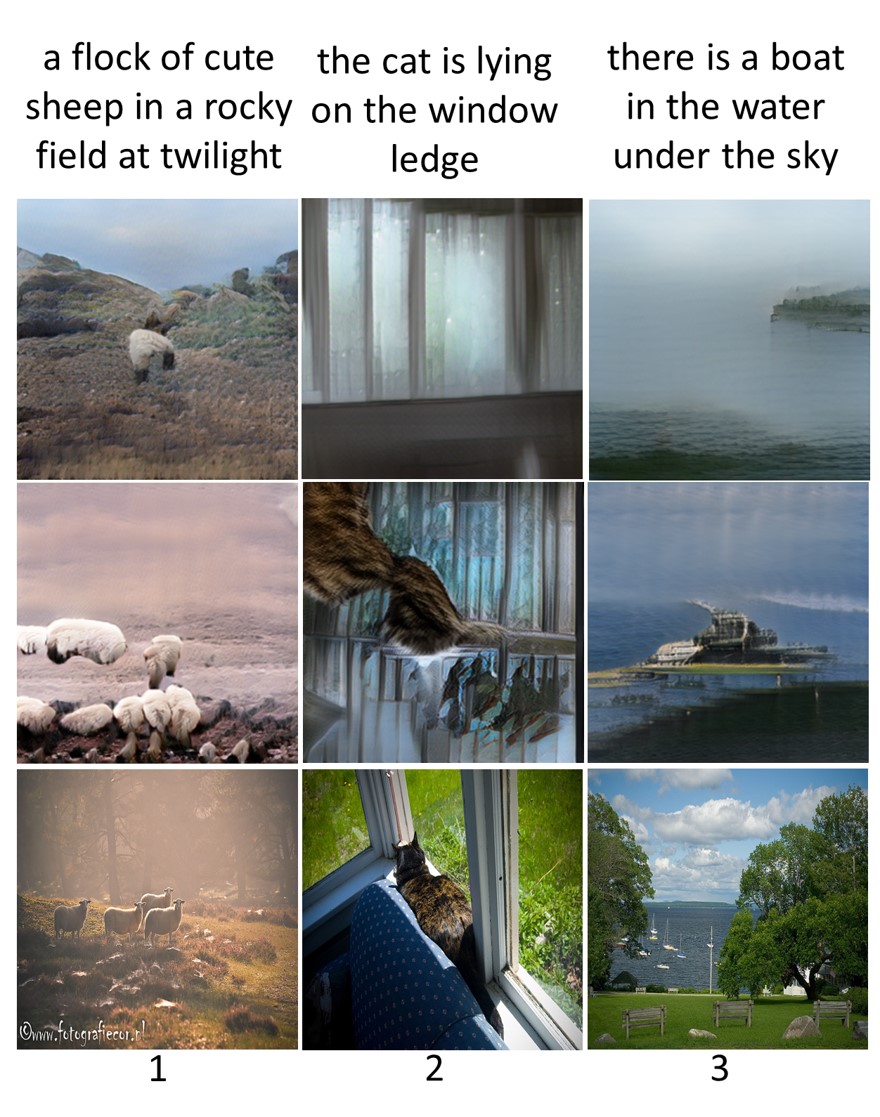}
}
\label{fig:comparision_b}
\subfigure[DMGAN VS DMGAN-VICTOR]{
\includegraphics[width=0.31\textwidth]{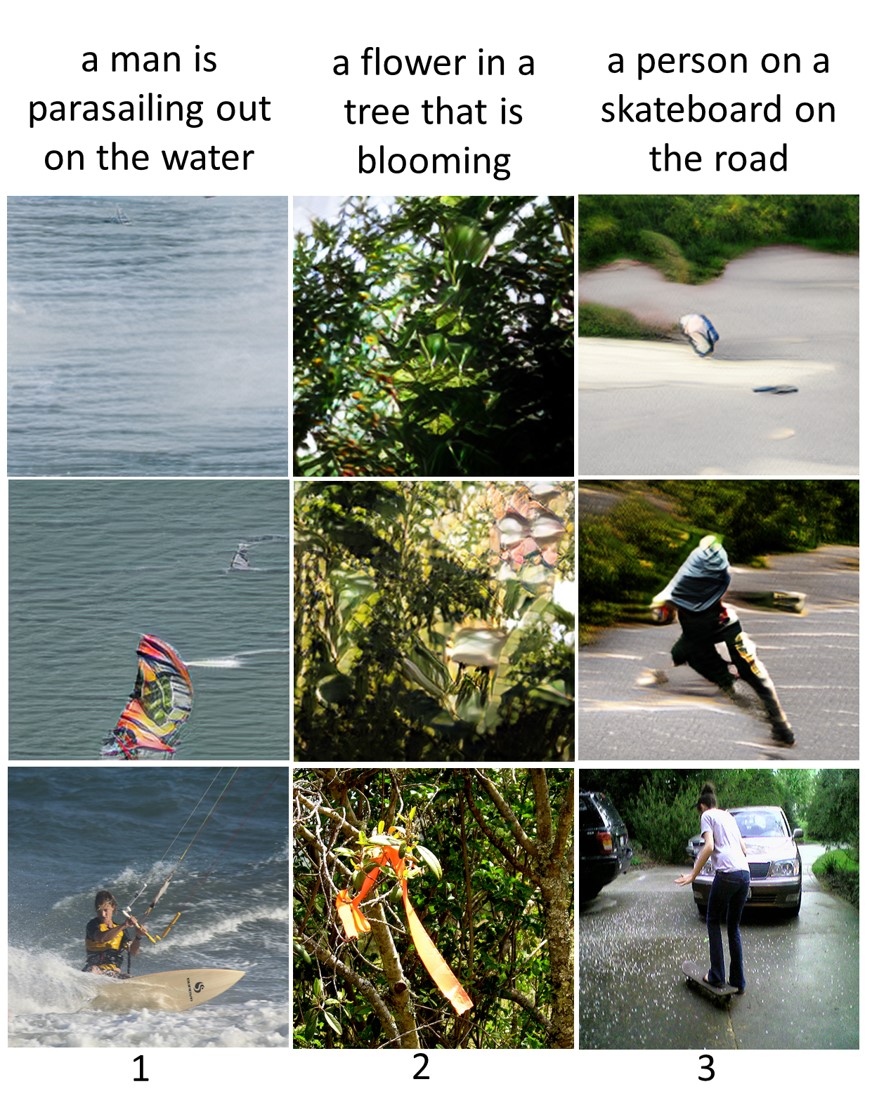}
}
\label{fig:comparision_c}
  \caption{Examples of images generated by (1st row) original StackGAN, AttnGAN, DM-GAN models, (2nd row) each model with VICTR, and (3rd row) the corresponding ground truth images.}
  \label{fig:comparison_stack}
  \vspace{-4mm}
\end{figure*}

\section{Evaluation Result}
\subsection{Quantitative Results}
Table \ref{tab:result_table} shows the performance of IS, FID and R-precision of the SOTA models, and the corresponding improvement with VICTR. Applying the VICTR-S feature in StackGAN improved the overall IS by around 1.93, which indicates the higher quality of final generated images. Specifically, the original StackGAN achieved 8.45 on IS with 600 epochs on stage-II, while the model with VICTR outperformed the original model at only 130 epochs. For AttnGAN and DM-GAN, we applied the VICTR-S feature in the initial image generation and VICTR-W feature for the iterative refinement. There is a clear improvement of all three metrics for both models with VICTR. The improvement in FID shows that using visual semantic relations between objects actually helped to form a group of objects in the geographically similar position to those in the ground truth images. Moreover, VICTR was mined from the original text and aligned the lingual semantics and visual semantics in the image captions. It helps the model to generate images which are better aligned to the text captions and leads to the increase of R-precision.

\begin{figure*}[t]
\centering
\subfigure[AttnGAN VS AttnGAN-VICTR]{
\includegraphics[width=0.48\textwidth]{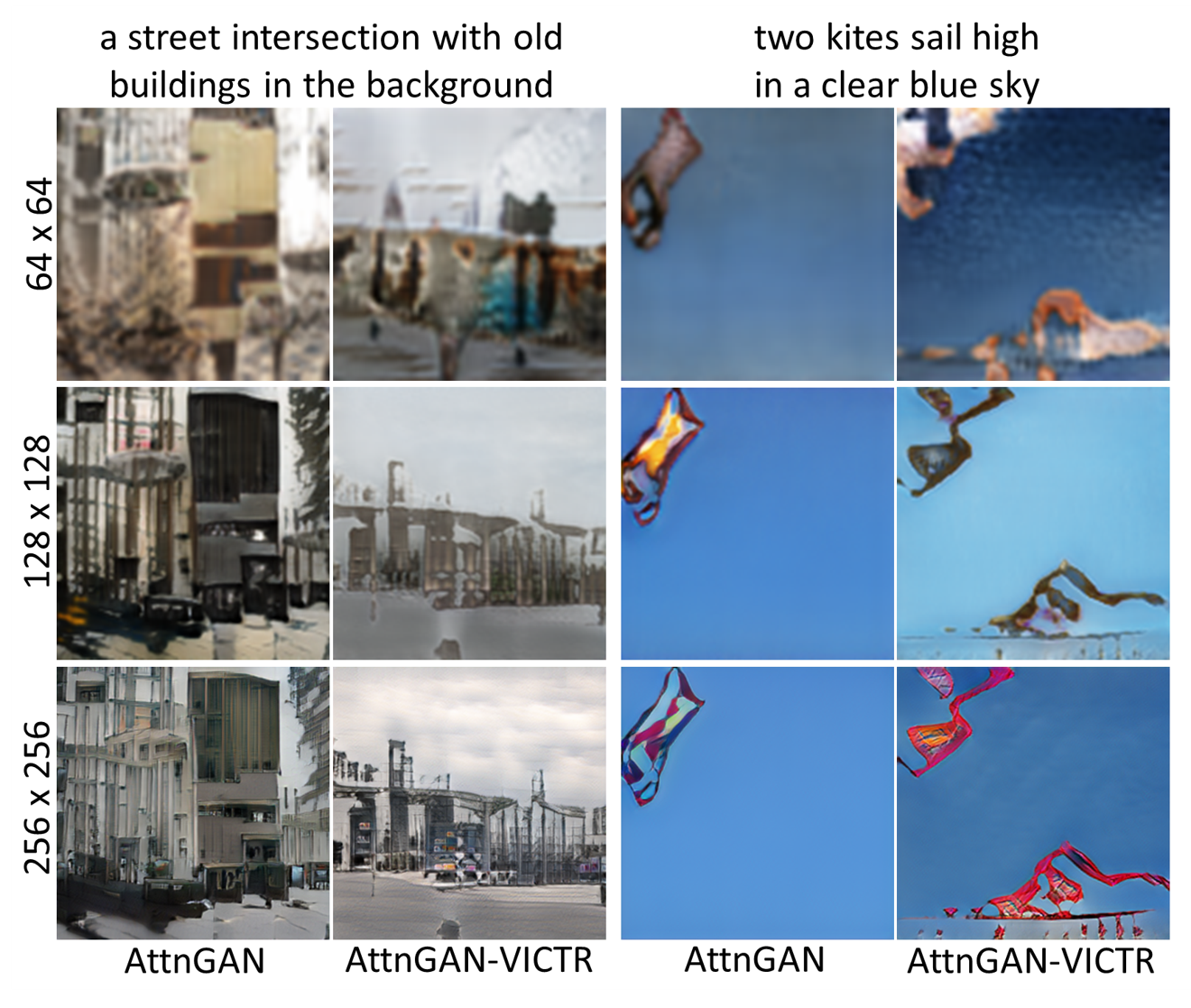}
}
\label{fig:3stage_a}
\subfigure[DMGAN VS DMGAN-VICTR]{
\includegraphics[width=0.48\textwidth]{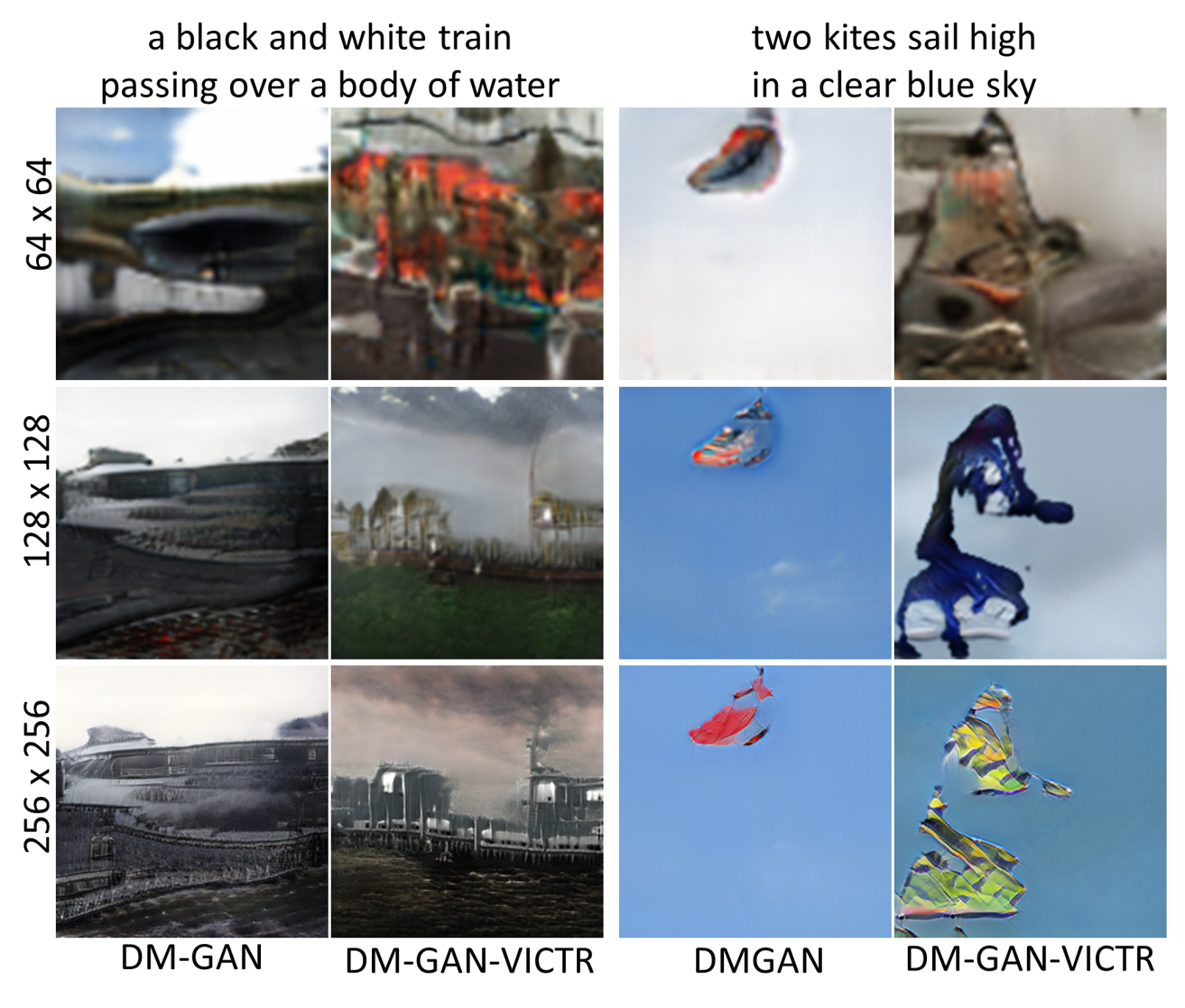}
}
\label{fig:3stage_b}
\caption[Optional caption for list of figures]{The result of 3 different stages of the original AttnGAN/DM-GAN and the AttnGAN-VICTR/DM-GAN-VICTR, including initial 64x64 image generation (1st row), the iterative refinement 128x28 images (2nd row) and 256x256 images (3rd row).}
\label{fig:3stage}
\label{fig:DM_3stage}
\end{figure*}

\subsection{Visual Comparison}
The visual comparisons between three SOTA baselines, and those with the proposed VICTR are presented in Figure \ref{fig:comparison_stack}. There are several findings from the visual comparison: firstly, with VICTR, images show a clearer structure (\textit{appearance of objects and their relative positions}) and are also closer to the ground-truth than those generated by the original SOTA model. For example, the column 3 in Figure \ref{fig:comparison_stack}(a) generated by StackGAN with VICTR has the similar structure of ground truth image that each object \textit{beach, ocean,} and \textit{sky} are positioned from the bottom to top, as well as a kite flying in the sky and a man standing in the beach. Secondly, compared to images from original models, the VICTR-based images provide clearer object shapes so the objects are relatively easier to be recognised, (i.e. \textit{food, plate, sheep, cat and human} from column 1,2 in Figure \ref{fig:comparison_stack}(a), column 1,2 in Figure \ref{fig:comparison_stack}(b) and column 1 in Figure \ref{fig:comparison_stack}(c)). Moreover, VICTR supports the model to well-understand contents in the text caption: 1) more objects from the text are identified in the image (e.g. the object \textit{cat} from column 2 in Figure \ref{fig:comparison_stack}(b) and the \textit{man/parasail} from column 1 in Figure \ref{fig:comparison_stack}(c) are completely missing without VICTR). 2) VICTR is good at handling quantifiers into individual objects. \textit{a flock of} sheep are well captured by VICTR at column 1 in Figure \ref{fig:comparison_stack}(b) where the original model failed to identify the number of objects. 3) even when the ground-truth image does not match with the caption, the VICTR-based models can generate images that are consistent with the caption, shown in column 3 and 2 in in Figure \ref{fig:comparison_stack}(b) and (c) respectively.

\begin{figure*}[]
\centering
  \subfigure[AttnGAN VS AttnGAN-VICTR]{
\includegraphics[width=0.48\textwidth]{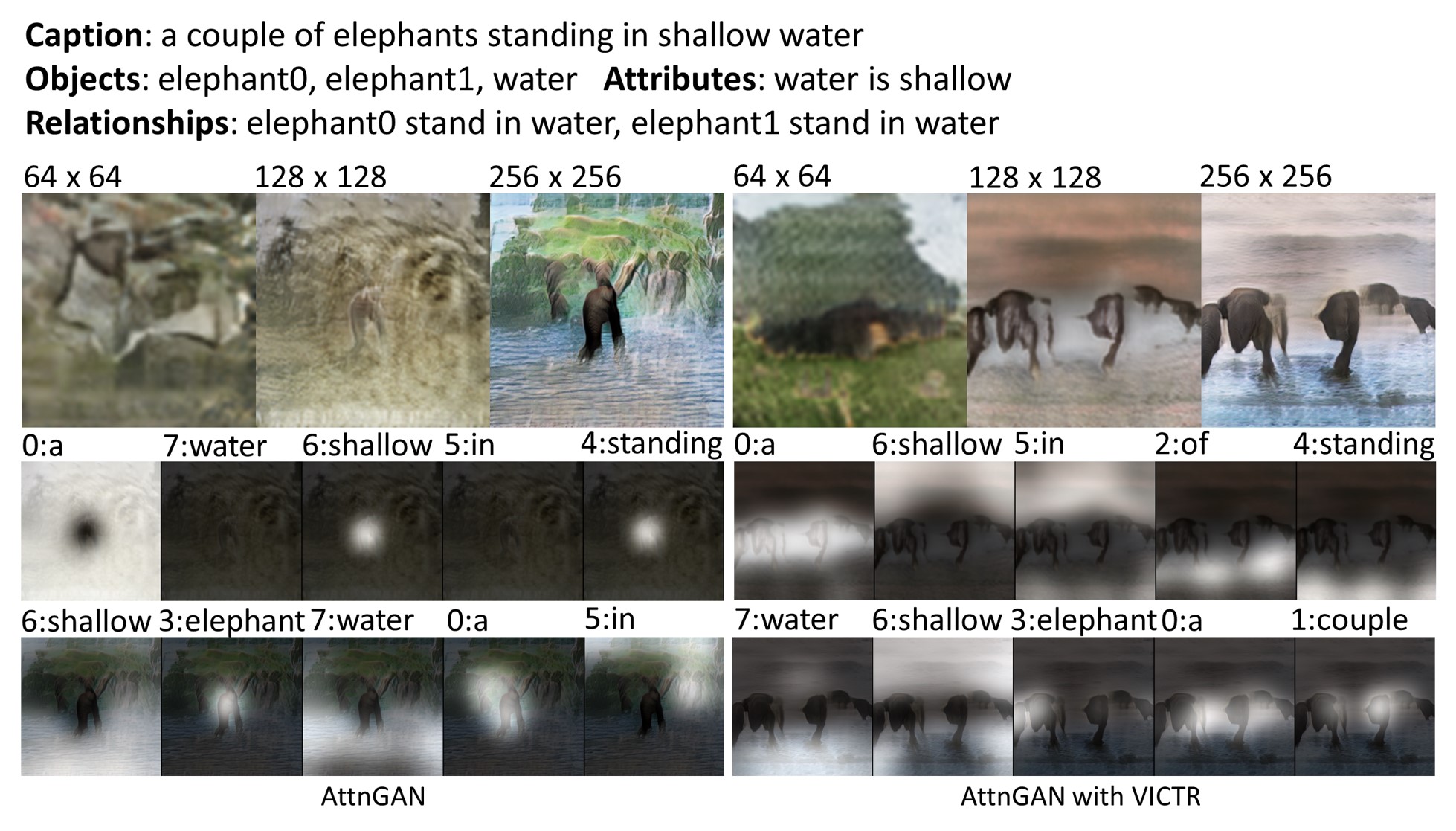}
}
\label{fig:3stage_a}
\subfigure[DMGAN VS DMGAN-VICTR]{
\includegraphics[width=0.48\textwidth]{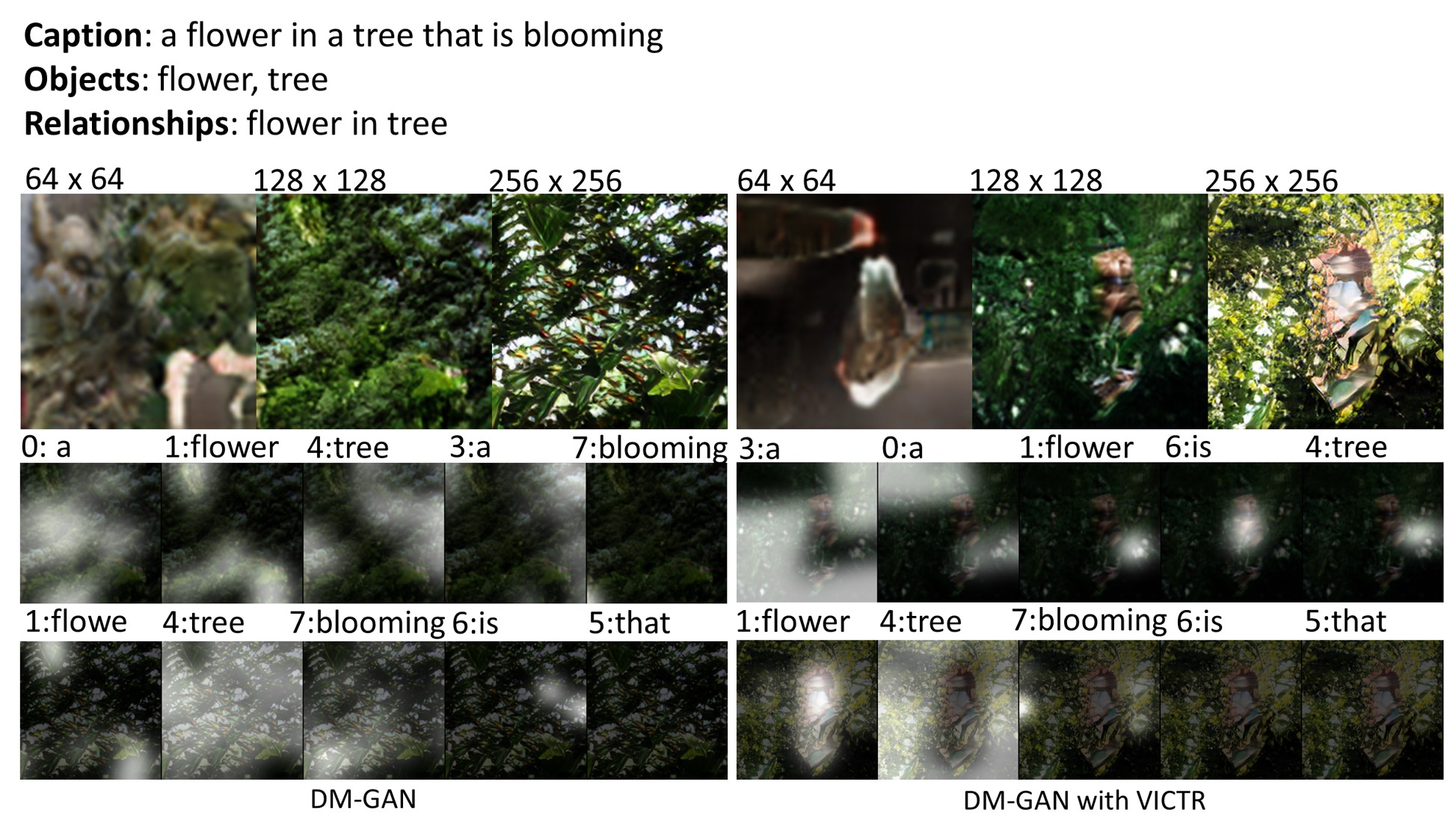}
}
\label{fig:3stage_b}
  \caption{Parsed Scene Graphs and Attention visualisation on the COCO2014. The first row shows the output 64x64, 128x128, and 256x256 images. The following rows show the attention map generated in stage 1 and 2 by the original AttnGAN/DM-GAN and with VICTR.}
  \label{fig:ATTN_attnmap}
\end{figure*}

\subsection{Ablation Study - Cascaded Generators}

Figure \ref{fig:3stage} indicates that VICTR-based model is able to generate better initialised images and refine them to be more related to the given text caption. In the baseline models, the initial stage image generation with the sentence-level feature captures the major frame or very rough appearance of objects identified from the text, whereas the image refinement process only focuses on the word-feature to polish the initial image but makes no major scene changes. From the images generated by original models, we found that: (1) the sentence-level feature from the Bi-LSTM encoder at the initial stage is not enough to produce the precise main image structure as described in the original text caption, so the models tend to create mistakes; and (2) the word feature in the following refinement process is not enough to amend these mistakes (from the initial stage), which limits the quality of final images. For example, the 2nd caption in Figure \ref{fig:3stage}(a) and \ref{fig:3stage}(b) describes \textit{two} kites sailing in the sky. However, both the initial images generated by the original AttnGAN and DM-GAN capture only \textit{one} kite in the sky and this error propagates to the final image. In comparison, in the images generated with VICTR, \textit{two} kites appeared from the initial stages, which matches the caption well and this persists all the way to the final image generation. Similar pattern can be found from the 1st caption in Figure \ref{fig:3stage}(a) where the original AttnGAN fails to well capture the positional relationship \textit{in the background} between the object \textit{street} and \textit{buildings} as well as the 1st caption in Figure \ref{fig:3stage}(b) from which the object \textit{train} and \textit{water} are not drawn clearly and not well positioned in relation to each other in the initial image, leading to the low quality of final image.

\subsection{Refinement Attention Inspection}
We visualise the parsed scene graph, and the intermediate images and attention maps of each refinement stage in Figure \ref{fig:ATTN_attnmap}. Several improvements can be observed in the word-image attention that better reflects the visual-linguistic alignment of objects and their positions: 1) The model with VICTR can focus on the more relevant and important object region in the image while using the corresponding word feature for the refinement. For example, in Figure \ref{fig:ATTN_attnmap}(a), the model with VICTR highlights the words \textit{a}, \textit{couple} and \textit{elephant} to generate \textit{two elephants} in the image whereas the original models do not. The similar pattern can be found with the words \textit{flower} and \textit{tree} in Figure \ref{fig:ATTN_attnmap}(b). 2) The positional relation attention represents a semantically meaningful visual context alignment on the linguistic relation expressed in the text description. This can be easily observed by the attention of words \textit{standing} \textit{in} from Figure \ref{fig:ATTN_attnmap}(a).

\subsection{Human Evaluation}
We conducted a human evaluation with 50 participants to qualitatively evaluate VICTR in the consistency between generated images and captions. The results and examples are in the Appendix \footnote{Available at: \url{https://usydnlp.info/victr_coling2020_appendix/}}.

\section{Conclusion}
In this paper, we proposed a new visual contextual text representation for text-to-image multimodal tasks, called VICTR, which extracts rich visual semantic information from input text descriptions. We have shown improvement across both quantitative and qualitative aspects when applying VICTR to diverse SOTA models in text-to-image generation. We also present an analysis showing the ability of VICTR to automatically organise different aspects of objects and learn the relations between them. The human evaluation results show that VICTR produces images that are highly aligned with text captions and very realistic. It is hoped that VICTR provides the insight into future integration of text handling in text-to-vision multimodal tasks.

\bibliographystyle{coling}
\bibliography{coling2020}

\end{document}